# End to End Recognition System for Recognizing Offline Unconstrained Vietnamese Handwriting

Anh Duc Le
*Center for Open Data in the Humanities*
Tokyo, Japan

Hung Tuan Nguyen and Masaki Nakagawa
*Department of Computer and Information Sciences*
*Tokyo University of Agriculture and Technology*
Tokyo, Japan

*Abstract*—Inspired by recent successes in neural machine translation and image caption generation, we present an attention based encoder decoder model (AED) to recognize Vietnamese Handwritten Text. The model composes of two parts: a DenseNet for extracting invariant features, and a Long Short-Term Memory network (LSTM) with an attention model incorporated for generating output text (LSTM decoder), which are connected from the CNN part to the attention model. The input of the CNN part is a handwritten text image and the target of the LSTM decoder is the corresponding text of the input image. Our model is trained end-to-end to predict the text from a given input image since all the parts are differential components. In the experiment section, we evaluate our proposed AED model on the VNOnDB-Word and VNOnDB-Line datasets to verify its efficiency. The experiential results show that our model achieves 12.30% of word error rate without using any language model. This result is competitive with the handwriting recognition system provided by Google in the Vietnamese Online Handwritten Text Recognition competition.

## I. INTRODUCTION

Recent years, the demand for reliable handwriting recognition systems has become more essential due to the development of pen-based/touch-based devices. These handwriting recognition systems should be able to recognize, retrieve, and search for handwritten text from digital ink (a sequence of pen-tip/finger-top coordinates) to meet a large number of requirements by users in practice. Thus, there are several studies in recent years, especially to focus on recognizing unconstrained handwritten text, which is written in each writer's style as usual without any constraint. The unconstrained handwritten text, so that, consists of a large number of variations in size, slant, skew, and stroke order. Although this problem has been studied for nearly 40 years, it is still a challenging topic in document analysis at present. For Western, Asian (Chinese/Japanese) and Arabic language, there are several studies on handwriting recognition including database acquisition, recognition, word spotting, and so on [1]–[5]. Recently, there are several impressive achievements on online unconstrained handwriting recognition such as English [6], Japanese [7], Chinese [8], [9], Arabic [10], and even mathematical formulae [11], [12].

During the last decade, the Vietnamese online handwriting recognition task has been raised and focused by a small number of studies [13]–[16]. Since these recognizers require pre-segmented handwritten text, they are not able to deal with the unconstrained handwritten text which are written cursively in practice. These studies mentioned the diacritical marks (DM) as the specific characteristic of Vietnamese script and proposed some solutions for solving the DM problem in character level. For unconstrained handwritten text, the position and order of DM strokes vary among different writers or even among different writing time of the same writer.

From 2008, the preliminary researches on isolated Vietnamese handwritten characters have been done so far. D. K. Nguyen and T. D. Bui have proposed a hierarchical algorithm to recognize online isolated Vietnamese characters, where the main characters and their DM are separately classified [13], [14]. They first employed Optimized Cosine Descriptor to represent a down-sampled multi-stroke handwriting pattern by a single vector (MOCD). The extracted MOCD vectors were fed into the main character classifier, which recognizes 26 categories (26 alphabet characters). Then, the circumflex DM classifier was employed to recognize the following DM: -, ^ and ?. Next, the tonal DM classifier was used for the other DM classification. In [13], [14], these three classifiers were Multi-Layer Perceptron (MLP) Networks and Support Vector Machines (SVM). For experiments, the authors combined their own dataset of 16,380 samples and Section 1c of Unipen dataset of 61,351 samples [17]. Their private dataset consisted of online handwriting patterns from 60 writers, who wrote 91 characters three times including vowels with DM. The average recognition rates of main character classifiers are 87.79% and 93.70% by MLP and SVM, respectively. For both circumflex and tonal DM classifiers, the average recognition rates are 93.88% and 94.66% by MLP and SVM, respectively.

D. C. Tran has proposed another approach to solve the problem with DM by segmenting the DM and main character before classifying them [15]. The key component was a segmentation algorithm to separate multiple strokes of a handwritten pattern into two groups: one being the main character and another being DM. Then, two separated classifiers were employed to recognize the main character and the DM, respectively. Final recognition result was obtained by combining and getting the highest result from two classifiers. The author employed both online and offline feature extractions. Both main character and DM classifiers were SVM. For experiments, the author employed the proposed method on both Unipen and IRONOFF datasets [17], [18] as well as his private dataset. The private dataset consisted of handwriting patterns from 14 combinations of DM (300 samples per combination), 25 characters without DM (400 samples per character), and 64 characters with DM (400 samples per character). For evaluating performance on separated tasks, the proposed method achieved 99.3% on DM classification and 95.2% on main character classification. For character recognition, the proposed method achieved 83.6% when the segmentation was not employed and 91.3% when the segmentation was used. The limitation of the preliminary studies as well as the reason for only a few handwriting recognition studies made in Vietnamese language was the lack of a benchmark database to compare different methods fairly. the authors usually presented the recognition results on their own private databases, which were difficult to verify or compare by others. Since 2016, a large Vietnamese Online Handwriting Database (VNOnDB) has been published and freely availed for research purpose [16]. In 2018, the Vietnamese Online Handwritten Text Recognition (VOHTR) competition was organized using VNOnDB [19], where several competitors published their works to be evaluated on the same database. The results from this competition can be considered as a good benchmark for recognition systems to be developed on Vietnamese Handwritten Text Recognition. In addition, the preliminary researches were not able to deal with the unconstrained handwritten text which is written cursively in practice, since these recognizers require pre-segmented handwritten text. In [14], the authors also proposed an unconstrained handwriting recognition method based on Bidirectional Long Short-Term Memory (BLSTM) and Connectionist Temporal Classification (CTC), which did not require any pre-segmentation information. They extracted point-based online features as well as local offline features. The best recognition accuracy was 92.32% on VNOnDB-Paragraph and 92.83% on VNOnDB-Line by two-layer BLSTM networks.

To overcome the pre-segmentation requirement of the preceding studies, we present an attention based encoder-decoder model to recognize unconstrained Vietnamese handwritten text at word level motivated by recent successes in deep learning. This paper is an extended and updated version of the previous conference paper [20] with a more elaborate and improvement of the architecture of ADE model. The standard CNN encoder and row BLSTM encoder [20] is replaced by DenseNet [24], a better feature extraction. Our system has two parts: a Convolution Neural Network (CNN) for extracting features, and a Long Short Term-Memory (LSTM) with attention mechanism as a decoder for generating output text named as LSTM decoder.

We do experiments on the VNOnDB database so that we could compare the performance of our system with the other preceding recognition systems. As the VNOnDB database composes of online handwritten text patterns, we convert these online handwriting patterns into offline patterns, which also eliminates the various orders of DM strokes written in the online patterns.

In the rest of this paper, the details of our system is presented in Section II. Then, the experiments and results is shown in Section III. Finally, our conclusion and discussion is given in Section IV.

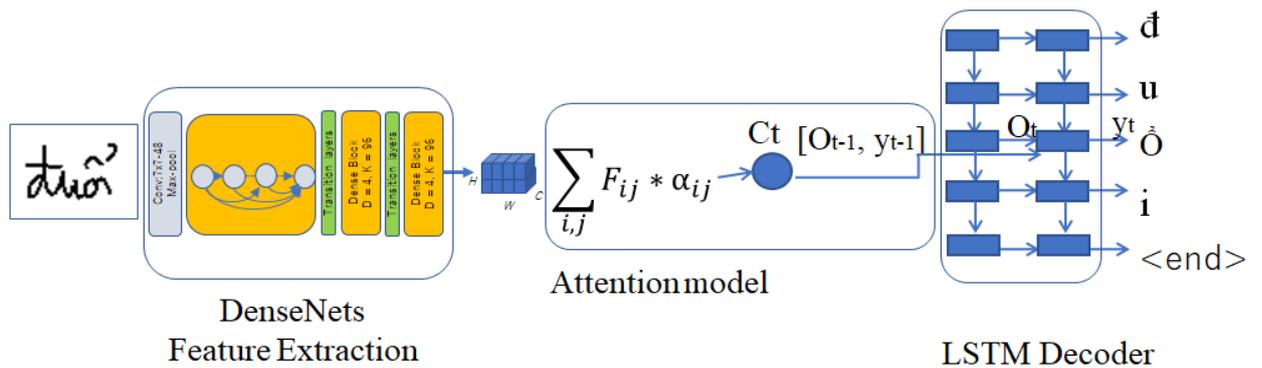

Fig. 1. Structure of the attention based encoder decoder model.

## II. METHODOLOGY

The structure of the attention based encoder decoder (AED) model is shown in Fig. 1. We replace the standard CNN encoder and row BLSTM encoder in our previous conference paper by DenseNet encoder, since the DenseNet has better performance than standard CNN for feature extraction. This is similar to the end-to-end recognition system for handwritten mathematical expression [21] – [23]. However, Vietnamese handwritings are handled as one dimension while handwritten mathematical expressions are handled as two dimensions. It has two main parts: DenseNet for extracting features from an image of handwriting, and a LSTM integrated with an attention model for generating the output text. Their details are described in the following subsections.

### A. DenseNet feature extractor

Features are extracted from an image of handwriting by a DenseNet which contains three densely blocks. Figure 2 shows the architecture of the DenseNet feature extraction. DenseNet outperforms the VGG and ResNet by proposing direct connections from any preceding layers to succeeding layers [23], [24]. The $i$th layer receives the feature maps of all preceding layers, $x_0, \ldots, x_{i-1}$, as input:

$$x_i = H_i([x_0, x_1, \ldots, x_{i-1}]) \qquad (1)$$

where refers to the convolutional function of the $i$th layer and refers to the concatenation of the output of all preceding layers. Densely connections help the network reuse and learn features cross layers. In each dense block, we add a blottleneck layers (1x1 convolution layer) before the 3x3 convolution layer to reduce the computational complexity. The dense blocks are connected by transition layers which contain convolutional and average pooling layers. For compression, the transition layer reduces a haft of feature maps. The detailed implementation is described as follows. We employ a convolutional layer with 48 feature maps and a max pooling layer to process input image. Then, we employ three dense blocks of growth rate (output feature map of each convolutional layer) $k = 96$ and the depth (number of convolutional layers in each dense block) $D = 4$ to extract features as Fig. 2. The size of the output features is $HxWxC$, where $H$, $W$, and $C$ are the height, width, and the depth of the extracted features.

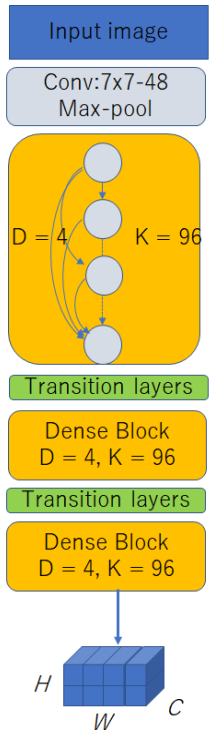

Fig. 2. Structure of the DenseNet feature extraction.

*B. Attention based LSTM decoder*

A decoder outputs one character at a certain time step. At each time step $t$, the decoder predicts symbol $y_t$ based on the current output $O_t$, and the context vector $C_t$. $O_t$ is calculated from the previous hidden state of the decoder $h_{t-1}$, the previous decoded vector $O_{t-1}$, and the previous embedded vector of the symbol $y_{t-1}$. $C_t$ is computed by weighted sum of the sequence of outputs and their weights produced by an attention model. The decoder is initialized by averaging the output of the encoder. The size of decoder LSTM is set as 256.

III. EXPERIMENTS

We trained the attention based encoder decoder model on the VNOnDB database. We set the learning rate as $10^{-8}$. The learning is decreased 10 times when the recognition rate on the validation set did not improve after 15 epochs. The training process was stopped when the learning rate reaches $10^{-11}$. After selecting the best model using the validation set, we evaluate it on the testing set and present the results in this section.

*A. VNOnDB dataset*

VNOnDB is collected from 200 writers who are asked to copy the ground truth text which are extracted from VieTreeBank (VTB) corpus [25]. The database is provided firstly in paragraph level (VNOnDB-Paragraph) and then the line and word segmentations are employed to produce the two datasets on line and word levels (VNOnDB-Line and VNOnDB-Word, respectively).

In the VOHTR competition, all of three levels are presented as three separated research tasks of the competition. However, we only make experiments on the VNOnDB-Word dataset to reduce the training time. In future research, we will try on the VNOnDB-Line and VNOnDB-Paragraph datasets.

Table I and II present the statistics on the number of lines, words, strokes and characters in the VNOnDB-Word and VNOnDB-Line datasets. Fig. 3 shows some samples of the same words written by different writers where Fig. 3 (a) and (b) are samples of the words "thành" and "công", respectively. There are several cursive writing styles for the same word. In addition, the position of DM strokes is various as well. Sometimes, the DM strokes is cursively written with the character strokes. VNOnDB-Word has the great varieties due to a large number of writers.

Since our system requires the offline images, we convert all online handwritten text of the VNOnDB-Word dataset into the offline images. Every online points of each online handwriting stroke is mapped into the two dimensional space of image so that the distance between the online points is proportional to the distance between them on the offline image. The stroke width on the offline image is fixed by 2 pixels. In addition, we render the offline images without applying the anti-alias or smooth methods. These offline images are not preprocessed.

TABLE I. STATISTICS OF THE VNONDB-WORD DATASET.

|  | Training set | Validation set | Testing set |
|---|---|---|---|
| **Number of words** | 66,991 | 18,640 | 25,115 |
| **Number of strokes** | 272,320 | 82,166 | 105,973 |
| **Number of characters** | 222,586 | 62,370 | 83,499 |

TABLE II. STATISTICS OF THE VNONDB-LINE DATASET.

|  | Training set | Validation set | Testing set |
|---|---|---|---|
| **Number of lines** | 4,433 | 1,229 | 1,634 |
| **Number of strokes** | 284,642 | 86,079 | 110,013 |
| **Number of characters** | 298,212 | 83,806 | 112,769 |

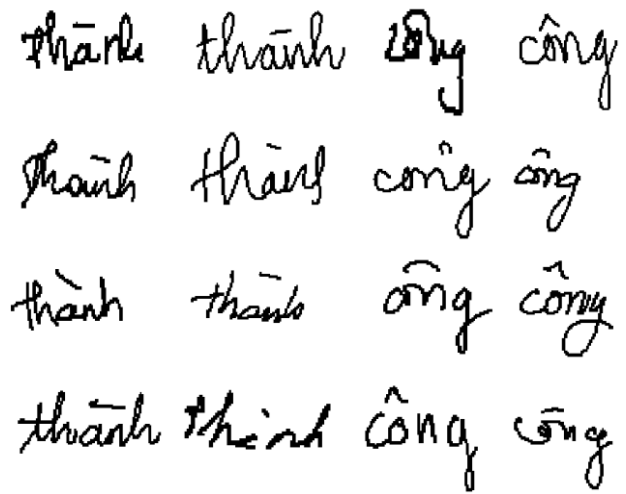

Fig. 3. Word examples in the VNOnDB-Word dataset.

## B. Evaluation metric

In order to measure the performance of our system on handwriting recognition, we use the Character Error Rate (CER) and Word Error Rate (WER) metrics which are generally employed for evaluating handwriting recognition systems. They are calculated based on the Normalized Edit Distance (NED) as the following equation:

$$NED(S_i, R_i) = \frac{100}{|S_i|} ED(S_i, R_i) \quad (2)$$

where $S_i$ is the *i*-th string belonging the set of target strings (ground truth text) S. $R_i$ is the corresponding output string of the $S_i$ string. $|S_i|$ is the number of words in $S_i$. ED is the edit distance function which computes the Levenshtein distance between two strings $S_i$ and $R_i$.

For CER, the ED is computed on the character level. For WER, the ED is computed on the word level. The CER and WER on the whole dataset was computed by averaging all pairs of $S_i$ and $R_i$.

## C. Results

Fig. 4 shows the CER and WER on the validation set during the training process of word level. The best CER on the validation set is . We employ the best model for running recognition on the testing set. For line level, we pretrained the recognition system by data of word level. Then, we finetuned the recognition system by data of line level.

Table III shows CER and WER of our recognition system and other participants of the VOHTR competition using VNOnDB-Word on the testing set. All of the competitors employed the different BLSTM network structures which were trained by CTC. The Google system consisted of several preprocessing techniques such as scaling, normalizing, resampling, and representing by Bezier curves, and so on. Then, the preprocessed handwritten patterns were passed through multiple BLSTM layers. The output of BLSTM layers were post-processed by a character n-gram language model, word n-gram language model and character class constraints.

For the IVTOV system, a line segmentation method was employed before applying preprocessing. The recognizer of this IVTOV system also used a BLSTM network with two BLSTM layers where each layer had 100 cells. The extracted features of the IVTOV system were delta of x-coordinates, delta of y-coordinates and pen up/down, which were simple online features. In the post-processing stage, the IVTOV system applied the dictionary constraints on the output of the BLSTM network.

Another submitted system in the competition came from MyScript. This system also employed several preprocessing techniques to normalize the digital ink, correct the slope and slant, etc. It composed of two recognizers, one is a feedforward neural network for predicting the single characters from the segmented candidates and another is a BLSTM network for predicting output text without segmentation. Since it worked with online handwritten patterns, the DM strokes were carefully processed to reduce the delayed strokes among the input strokes. For post-processing, this system used a syllable-based unigram language model from the VTB and other corpora. As shown in Table III, our recognition system achieved 4.10% of CER and 10.24% of WER on testing set. Although our model did not employ any language model as other participants, our results are better than our previous version and the GoogleTask1 system. We can improve CER and WER of our system with a language model. In addition, our system is designed and trained to recognize both offline and online handwriting text.

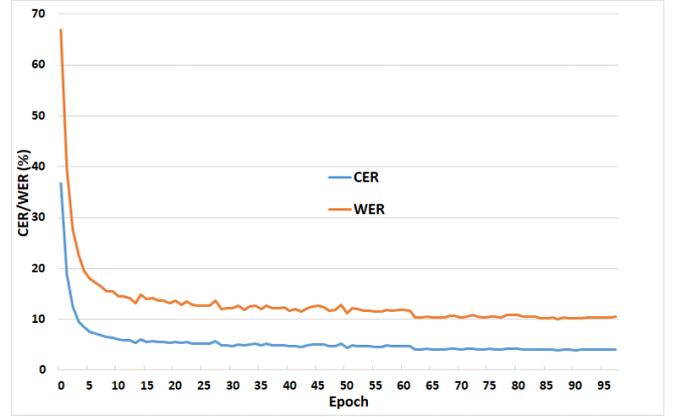

Fig. 4. The WER (%) on training, validation and testing sets during training process.

TABLE III. THE RESULTS OF RECOGNITION SYSTEMS ON THE VNONDB-WORD TESTING SET.

| System | Corpus | CER(%) | WER (%) |
|---|---|---|---|
| GoogleTask1 | Other | 6.09 | 13.18 |
| IVTOVTask1 | VTB | 2.92 | 6.47 |
| MyScriptTask1 | VTB + Others | 2.91 | 6.46 |
| Our model with standard CNN and BLSTM encoders | None | 5.88 | 12.30 |
| Our model with DenseNet encoder | None | 4.10 | 10.24 |

Table IV shows CER and WER of our recognition and other participants on the VNONDB-Line testing set. Our recognition achieved 4.67% of CER and 13.33% of WER. Our recognition system outperforms GoogleTask2 system on CER while it outperforms both the GoogleTask2 and IVTOVTask2 systems on WER.

TABLE IV. THE RESULTS OF RECOGNITION SYSTEMS ON THE VNONDB-WORD TESTING SET.

| System | Corpus | CER (%) | WER (%) |
|---|---|---|---|
| GoogleTask2 | Other | 6.86 | 19.00 |
| IVTOVTask2 | VTB | 3.24 | 14.11 |
| MyScriptTask2_1 | VTB | 1.02 | 2.02 |

| MyScriptTask2_2 | VTB + Others | 1.57 | 4.02 |
| Our model with DenseNet encoder | None | 4.67 | 13.33 |

Fig. 5 shows the recognition process of the attention based encoder decoder model. The model generates characters until it reaches the <end> symbol. At each time step, the decoder focuses on a part of the input image (red part in the image) to generate the corresponding character at the top of the image. We show some correctly recognized samples in Table V. The first column contains the handwritten words, the second column is ground truth text and the last column is the output of our system. In the second and last columns, we insert the spaces between the characters of a word to show every characters clearly.

On the other hand, Table VI shows misrecognized samples by our system. Note that, we insert the spaces between the characters of a word to show every characters clearly in the second and third columns. Among these cases, the two first rows of "L o n g" and "s ử a" are ambiguous even for people, which might depend on the writing styles of writers or be wrongly written. These two first outputs of our system are still acceptable with the input images. Our system can be fine-tuned for each writer while being deployed in practical, so that it is able to adapt itself to the writing style of the writer. In the third row, the error case of "T á m", which is predicted as "T h á m" might relate to the over attention problem. In the last row, the case of "h ứ a" and "h ừ a" could be modified by lexicon constraints such as a dictionary or language model.

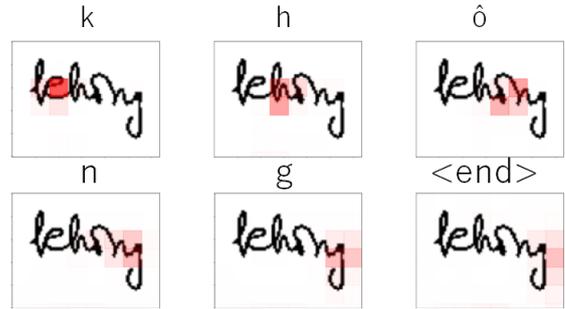

Fig. 5. Visualizing the The recognition process of attention based encoder decoder model.

Fig. 6 shows three examples of the recognition results of handwritten text lines. The recognition results are shown below the handwritten lines. The misrecognitions are shown in red color. Even handwritten inputs are cursive and very hard to read by human, our recognition provides precise results. The mis-recognitions are acceptable. We can employ a language model to revise those mis-recognitions.

TABLE V. CORRECTLY RECOGNIZED EXAMPLES.

| Input images | Ground truth | Recognition result |
|---|---|---|
| *Tôi* | t ô i | t ô i |
| *Từng* | t ừ n g | t ừ n g |
| *nghĩ* | n g h ĩ | n g h ĩ |
| *dành* | d à n h | d à n h |

TABLE VI. INCORRECTLY RECOGNIZED EXAMPLES.

| Input images | Ground truth | Recognition result |
|---|---|---|
| *long* | L o n g | l o n g |
| *sửa* | s ử a | s ứ a |
| *Tám* | T á m | T h á m |
| *hứa* | h ứ a | h ừ a |

tí Đồng Hứu ( Quảng Bình ) vẫn vào Quảng Trị thấy nhưng trong những nghĩa trang nơi

hàng chụu vạn người lính đã nằm xuống cho độc lậy tự do của Tổ quồi, anh cũng chỉ là

Chẳng hạn như trường hợp Phước " mì " bán vé số ở ấp Phước Hòa , xã Thành

Fig. 6. Example of the recognition results on VNOnDB-Line dataset.

## IV. CONCLUSION

In this paper, we have presented the attention based encoder decoder model (AED) for recognizing Vietnamese handwriting. The model is trained end-to-end with input handwritten images and target characters. We achieved 4.10% of Character Error Rate (CER) and 10.24% of Word Error Rate (WER) on the testing set of VNOnDB-Word and 4.67% of CER and 13.33% of WER on the testing set of VNOnDB-Line. It is shown that our model outperforms the GoogleTask1, GoogleTask2, and IVTOVTask2 systems while our model did not employ any language model. The AED model is a potential model for handwritten recognition. We plan to use the AED model for other languages such as Japanese, Chinese and integrate with language models in the future research.